\documentclass[conference]{vgtc}                
\ifpdf
  \pdfoutput=1\relax                   
  \usepackage{graphicx}                
  \DeclareGraphicsExtensions{.pdf,.png,.jpg,.jpeg} 
\else
  \usepackage{graphicx}                
  \DeclareGraphicsExtensions{.eps}     
\fi%
\graphicspath{{figures/}{pictures/}{images/}{./}} 

\PassOptionsToPackage{dvipsnames}{xcolor}
\vgtcinsertpkg

\usepackage{url}
\usepackage[utf8]{inputenc}
\usepackage[T1]{fontenc}
\usepackage{microtype}                 
\PassOptionsToPackage{warn}{textcomp}  
\usepackage{textcomp}                  
\usepackage{mathptmx}                  
\usepackage{times}                     
\usepackage{cite}                      
\usepackage{booktabs}                  
\usepackage{amsmath}
\usepackage{multirow}
\usepackage{cleveref}
\usepackage{t1enc}
\usepackage{tabularx}
\usepackage{makecell}
\usepackage{xspace}
\usepackage{soul}
\usepackage{color}
\usepackage{relsize}
\usepackage{enumitem}
\usepackage[plain]{algorithm}
\usepackage[noend]{algpseudocode}
\usepackage[labelfont=sf]{subcaption}
\usepackage{tcolorbox}

\onlineid{5164}

\vgtccategory{Research}
\vgtcpapertype{application/design study}

\title{Enabling Fast and Accurate Crowdsourced Annotation for Elevation-Aware Flood Extent Mapping}

\author{Landon Dyken\thanks{e-mail: ldyke@uic.edu}\\ %
        \scriptsize University of Illinois Chicago %
\and Saugat Adhikari\thanks{e-mail: adhiksa@iu.edu}\\ %
     \scriptsize Indiana University Bloomington %
\and Pravin Poudel\\ %
     \scriptsize Utah State University
     \and Steve Petruzza\\ %
     \scriptsize Utah State University
     \and Da Yan\\ %
     \scriptsize Indiana University Bloomington
     \and Will Usher\\ %
     \scriptsize Luminary Cloud
     \and Sidharth Kumar\thanks{e-mail: sidharth@uic.edu}\\ %
     \scriptsize University of Illinois Chicago}

\DeclareRobustCommand{\highlight}[1]{{#1}}

\abstract{%
Mapping the extent of flood events is a necessary and important aspect of disaster management. In recent years, deep learning methods have evolved as an effective tool to quickly label high-resolution imagery and provide necessary flood extent mappings. These methods, though, require large amounts of annotated training data to create models that are accurate and robust to new flooded imagery. 
In this work, we present FloodTrace, a web-based application that enables effective crowdsourcing of flooded region annotation for machine learning applications. To create this application, we conducted extensive interviews with domain experts to produce a set of formal requirements. Our work brings topological segmentation tools to the web and greatly improves annotation efficiency compared to the state-of-the-art. 
The user-friendliness of our solution allows researchers to outsource annotations to non-experts and utilize them to produce training data with equal quality to fully expert-labeled data.
We conducted a user study to confirm our application's effectiveness in which 266 graduate students annotated high-resolution aerial imagery from Hurricane Matthew in North Carolina. Experimental results show the efficiency benefits of our application for untrained users, with median annotation time less than half the state-of-the-art annotation method. In addition, using our application's aggregation and correction framework, flood detection models trained on crowdsourced annotations were able to achieve performance equal to models trained on fully expert-labeled annotations, while requiring a fraction of the expert's time. 

}

\teaser{
 \centering
 \includegraphics[width=0.95\linewidth]{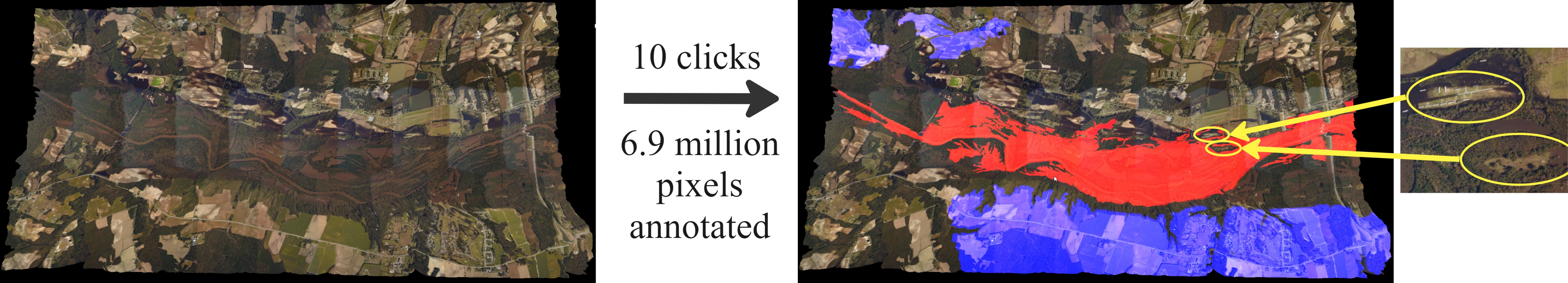}
 \vspace{-0.5em}
  \caption{\label{fig:teaser}%
  An annotation made on a 6500 $\times$ 3000 pixel aerial image of flooding caused by Hurricane Matthew using only ten clicks of our topological segmentation tool. Pixels overlaid with blue have been labeled as dry and red as flooded. Note that areas labeled as flooded can appear green on the left, this is due to the tree canopy obscuring the flooded ground beneath. This tool guides annotation using the corresponding elevation data for a region, resulting in accurate labels. This can be seen by the highlighted areas where the tool annotated flooding around small dry features, but left them unlabeled. In our experimental user study, we show that this tool is more efficient than the state-of-the-art elevation-guided tool and can be used effectively by untrained participants.
  }
}
\date{}

\keywords{Terrain visualization, machine learning, collaborative and distributed visualization, topology-based techniques}

\CCScatlist{ 
 \CCScat{Human-centered computing}{Visualization}{Visualization application domains}{Geographic visualization
};
}

\begin{document}

\firstsection{Introduction}

\maketitle

\label{sec:introduction}
In recent years, the societal impact of flooding has been difficult to ignore, with the severity and frequency of flood events increasing year after year \cite{gar_2022}. In response, mapping the extent of these floods has become a crucial tool in a multitude of domains, from disaster response and insurance risk assessment to urban planning and agriculture. 

As flood extent mapping continues to grow in importance, the amount of data being gathered on flooded regions has also grown, with it now being unfeasible to expect manual labeling of flood extent to meet necessary demand. To address this, it has become common for much smaller amounts of labeled data to be used to train classification models which can then quickly perform this task on new data as it is collected. Many classification models based on deep learning techniques have shown effectiveness for flood extent mapping by utilizing satellite imagery, unmanned aerial vehicle (UAV) data, hydrographs, and digital elevation model (DEM) data \cite{floodsurvey_2022}.

Although these models greatly reduce the area of flooded regions that need to be manually labeled, they still require annotated training sets to effectively learn from. High accuracy within training sets is paramount to allow dependent models to correctly learn and predict flooding. Producing these annotations requires a huge time commitment for domain scientists, which becomes prohibitive for creating the large, varied datasets necessary to train models that can reliably respond to unseen flooded imagery. While fully expert-labeled training data is the standard, crowdsourcing offers a powerful way to gather annotated datasets, and has been shown before as a valuable tool for emergency response \cite{gao_crowd_2011, goodchild_crowd_2010} and mapping floods specifically \cite{liang_flood_2018, degrossi2014flood, sunkara2020street}.  

\highlight{In this domain, annotation consists of assigning categories (such as flooded or dry) to the pixels of an image, usually through directly applying color.} Existing flood annotation datasets are most commonly created by manual labeling of aerial imagery with simple brushes and polygon selection tools, which is extremely time-consuming. Semi-automatic approaches improve a user's productivity by allowing them to label larger areas more quickly with the assistance of a guiding algorithm. Semi-automatic tools have been used previously to create flood annotation datasets \cite{gebrehiwot_flood_2019, chowdhury_rescuenet_2022, rescuenet_2021, liang_flood_2018} through various algorithms on the imagery being labeled. While these tools can improve efficiency, our work builds on insight from domain experts which motivates a focus on \textit{elevation-guided} annotation of flooded regions. 
Recently, elevation guided methods have become more common, where satellite imagery is supplemented with a corresponding elevation model to improve the accuracy of flood extent mapping. The state-of-the-art machine learning models developed for this task are elevation-guided and use physics-informed algorithms that utilize Digital Elevation Model (DEM) data \cite{xie_markov_2018, jiang_markov_2019, hashemi-beni_region_2021, munoz_df_2021, sainju_markov_2022}. These models require very accurate ground truth labels for training. This is especially true for areas of high and low relative elevation in a dataset, and for adjacent pixels with different elevation, named border pixels. With these models, a high-elevation pixel that is mislabeled as flooded can cause many surrounding pixels to be incorrectly inferred as flooded. By using both aerial imagery and corresponding elevation data, elevation-guided annotation can create more accurate training data for dependent models. 

In this paper, we present FloodTrace, an interactive web-based system for quick, accurate annotation that improves researchers' workflows and enables high-quality crowdsourced labeling for flood extent mapping. Our design was informed by domain experts through several meetings and demos with one senior and two junior researchers working on machine learning models for flood extent mapping. In our system, we take advantage of elevation data to inform annotation decisions with 3D interactive visualization and semi-automatic annotation algorithms. 

FloodTrace is designed to encourage and improve the quality of crowdsourcing for this domain. For this goal, our solution is implemented as an interactive web application, making it easily accessible for end users. Our systems brings for the first time topological data analysis to a convenient web environment; it provides the ability to select features based on topological segmentations, a process that would otherwise require complex setup, expertise with topological toolkits, and slow, exhaustive workflows. To improve researcher trust and understanding, our application provides an uncertainty visualization component to visualize aggregated crowdsourced annotations and find areas of high uncertainty between annotators. Our annotation tools can be used to directly correct labels for these uncertain areas, quickly improving the quality of the aggregated annotations.

To evaluate FloodTrace, we conducted a user study with 266 computer science graduate students in which we logged all interactions and results over several annotation tasks. Aggregated annotations were then used to train elevation-guided machine learning models, which were tested for flood detection on unseen regions to assess training data quality. In our experimentation, we found that elevation-guided tools increased the accuracy of participants' annotations, while our topological segmentation tool greatly increased annotation efficiency. We show that by using our aggregate visualization and correction workflow, crowdsourced annotations can be used to create models with equal performance to models trained on fully expert-labeled annotations, while requiring a fraction of an expert's time. In summary, our contributions are:

\begin{itemize}[leftmargin=*]
\itemsep0em
\item \highlight{An intuitive web-based tool for crowdsourced annotation of flood inundation maps that leverages satellite imagery, elevation data, and topological segmentation to enable non-experts to quickly and accurately label flooded areas. Our application is the first to bring users access to topological segmentation techniques on the web.}
    \item  \highlight{A novel review tool for visualizing aggregated crowdsourced flood annotations that enables interactive correction to improve labels.}
    \item \highlight{An evaluation of our framework through a user study with 266 participants, producing a total of 1,321 unique annotations over eight distinct regions of North Carolina flood areas during Hurricane Matthew.}
\end{itemize}

\section{Background and Related Work}
\label{sec:related_work}
In this section, we first discuss the workflow of machine learning researchers to provide context for how our application can improve existing methods for flood extent mapping (\autoref{sec:workflow}). Then, we explore previous work in crowdsourcing applications to motivate and inform our design (\autoref{sec:crowdsourcing}) and review related work in annotation tools (\autoref{sec:prev_tools}). Finally, we motivate and give background for the topological data analysis methods used for our novel elevation-guided annotation tool (\autoref{sec:topology_analysis}).

\subsection{ML for Flood Extent Mapping Workflow}
\label{sec:workflow}
While workflows can vary based on the specific application, we report here an example workflow of training models for flood extent mapping from a discussion with our expert collaborators. 

\paragraph{Data acquisition and pre-processing:} The process begins with data collection, acquiring aerial imagery taken during a flood disaster with corresponding digital elevation model (DEM) data. This imagery can be obtained from a source such as NOAA's National Geodetic Survey of Emergency Response Imagery \cite{noaa} in the form of patches with global Coordinate Reference System (CRS) info. These patches are then stitched together into test regions by loading them into QGIS\cite{qgis}, finding which patches are needed to create a test region using their CRS info, and then using the GDAL\cite{gdal} library in Python to combine them into one image. DEM data can then be created for that region in the same way by downloading and stitching together data patches at corresponding CRS values, with DEM data widely available for locations in the US\cite{Gesch_ned_2018} and globally\cite{abrams_gdem_2020}. Afterward, both imagery and DEM data need to be resampled into the same spatial resolution so that the pixels are aligned. 

\paragraph{Image annotation, model training, and inferencing:} The next step is to annotate the flooded and dry areas in the imagery to obtain ground truth labels for model training and evaluation. Study areas for flood mapping can easily have millions of pixels, so while this can be done with semi-automatic tools, it is still the most time-consuming part of the flood extent mapping process, taking multiple hours per region. 
In order to create ML models that can robustly detect flooding on unseen data, large training sets are required. While existing work has explored using fully automated methods for creating labeled flood datasets\cite{Bonafilia_data_2020}, they are not widely used in this domain because of quality concerns and the added difficulty for researchers to understand and explain model behavior after training. In our collaborators' lab, the bulk of annotation is accomplished by trained graduate students. In our work, we hope to improve the workflows of ML researchers by reducing the burden of annotation through crowdsourcing and more efficient semi-automatic tools. 

\subsection{Crowdsourcing for Research}
\label{sec:crowdsourcing}
Many projects (e.g. eBird\cite{wood_ebird_2011}, GalaxyZoo\cite{lintott_galaxyzoo_2008}) have shown that crowdsourcing and citizen science can be successful for large-scale data collection and annotation. Because any barrier to entry will lower the number of users of a platform, crowdsourcing applications are almost always web-based for ease of access. To define the qualities of successful crowdsourcing projects, Law et al. \cite{law_crowdsourcing_2017} deduced a set of requirements that make crowdsourcing feasible, desirable, and useful for a given research problem. Flood extent mapping satisfies their challenges of data sensitivity, quantity, and availability, along with crowd interest, intention, and ability as proven by other applications around crowdsourced flood event data\cite{songchon_crowd_2023, panteras_crowd_2018, see_crowd_2019}. In addition, previous work\cite{liang_flood_2018} has shown that, while individual crowdworkers may not perform well in flood mapping, aggregating annotations can lead to high-quality results. This work aggregates user annotations by labeling patches as flooded or dry if the ratio of received labels for that patch is above an empirically chosen certainty threshold. While we experimented with this method of aggregating annotations, we found better ML model performance by training using scored soft labels of flood and dry created from averaging user annotations. Within crowdsourcing tools, FloodTrace is unique in providing a system for experts to quickly visualize areas of uncertainty between annotators and apply new labels to revise and improve the set of annotations. 

\subsection{Annotation Tools}
\label{sec:prev_tools}
Multiple libraries such as OpenStreetMap\cite{grinbergerOpenstreetmap2022}, QGIS Cloud\cite{qgis}, and ArcGIS Online\cite{arcgis} provide annotation tools for geospatial data. Each of these tools was made for different use cases but can support DEM data, interactive 3D rendering, and basic annotation with text, polygons, and brushes. Their potential for flood extent mapping, however, is limited by the fact that none provide semi-automatic annotation tools for the purpose. Even with crowdsourcing, manual annotation of flooded imagery with polygon selections and brushes is too time-consuming to be desirable and results in less accurate training data. 

For semi-automatic annotation, there are two main classes of tools in this domain. The first are those that produce a segmentation of the input and allow users to apply flood or dry class labels to selected segments (\textit{segmentation-type} tools). Liang et al. \cite{liang_flood_2018} use imagery patches exported from a graph-based clustering approach as these segments, and show that these patches can be labeled effectively by non-experts. Other work has used image features produced by class-agnostic neural networks for labeling to create large flood annotation datasets \cite{chowdhury_rescuenet_2022, rescuenet_2021}. The second class of tools are those where the user selects seed pixels which are then extended with connected pixels by some rule (\textit{extension-type} tools). As an example, Gebrehiwot et al. \cite{gebrehiwot_flood_2019} use a tool to automatically label connected pixels of the same color as flooded or dry. While the semi-automatic annotation tools mentioned so far operate directly on aerial imagery data, recent exploratory work has been done on creating an elevation-guided extension-type tool for labeling flooded datasets \cite{adhikari_elevation_2022}. This work utilizes DEMs to select all connected downstream (lower elevation) or upstream (higher elevation) pixels from a selected seed pixel, which they call the breadth-first search (BFS) method. We adapt their BFS method as a state-of-the-art tool in our application (\autoref{sec:BFS}). It is important to note that the original work\cite{adhikari_elevation_2022} does not support interactive 3D rendering, can only operate on small data patches, and requires installation, all issues that are addressed by our web-based solution with GPU accelerated rendering.

\subsection{Topological Analysis of Elevation Data}
\label{sec:topology_analysis}
Topological methods provide powerful tools for analyzing elevation data features that are critical for flood mapping. Previously, they have been successfully used to simplify and extract features of interest in a variety of scientific domains \cite{bhatia2018topoms, bock2017topoangler, petruzza2019high, gyulassyGeometry2016, rieckNetworks2018} and specifically applied to DEM data. Yu et al. \cite{yusimplification2021} propose a method combining structural analysis and statistical filtering for terrain simplification while preserving smooth morphology and structural edges. Guilbert \cite{guilbertfeatures2013} uses contours to generate a feature tree for underlying data at multiple levels of detail. Wu et al. \cite{wucontourtree2015} use a localized contour tree method to detect and characterize surface depressions across scales. Recently, Corcoran et al. \cite{corcoran2023topological} show the advantages of persistent homology methods in terrain analysis for their robustness to noise and facilitating of machine learning methods.

The challenge in analyzing elevation data lies in its multi-scale nature, with terrain features existing at widely varying scales. This makes it difficult to identify relevant features, making multi-scale views required in order to conduct meaningful analysis \cite{sunsimplification2018, yusimplification2021, wucontourtree2015, guilbertfeatures2013, feciskaninsimplification2021}. To address this challenge, we utilize persistent homology, a mathematical framework that quantifies the importance of topological features across different scales. Persistent homology is a powerful tool to remove topological features smaller than a given scale \cite{edelsbrunner2000topological}. This is done by considering the persistence of these features, which is their lifespan when conducting a sweep through the range of the function. Simplification by persistence involves removing a topological feature if its lifespan is below some threshold. In doing this, simplification by persistence is robust for maintaining topological features in high-resolution data and preserves salient details. As an example, a spatially small feature with a huge variance in elevation compared to its surroundings would have a large persistence value, and not be simplified until high thresholds, whereas downsampling the data's spatial resolution would quickly remove this feature.

Along with persistent homology, we utilize contour tree segmentation to create a hierarchical representation of the elevation data. We experimented with topological segmentations produced by the Morse-Smale complex (MSC) \cite{edelsbrunnerMorse2003}, merge, split, and contour trees, and found the contour tree to produce segments most closely aligned with flooded and dry regions while respecting elevation border pixels. Contour trees segment data by considering contours through a sweep of the function range, defining segments as the birth or split of a contour until it merges with another or vanishes. When simplifying by persistence, less persistent topological features are collapsed to the same elevation level as more persistent neighboring features. Because contour tree segments are regions bounded by contours at a given elevation level, this simplification corresponds to increasing the size of segments by collapsing the neighboring less persistent features into them. By considering multiple levels of simplification, this gives a convenient segmentation for selecting potential water levels in a flooded region, as water will naturally pool into depression features and stop at hill features or elevation contours.

Similar topological techniques have previously been utilized for semi-automatic annotation in neuron tracing of large brain volumetric data \cite{mcdonald2020improving}. This work utilizes the MSC to extract ridge-like structures which correspond to neuron center-lines, which can then guide user tracing. Simplification by persistence is used with a single empirically chosen persistence threshold to remove noise and create a sparser dataset for users to interact with. While taking inspiration from this work, our system adapts topological methods specifically to the labeling of flood extent in terrain data using contour tree segmentations. 
Furthermore, our system allows interactive selection of multiscale features by providing data simplified at varying persistence thresholds. This is 
necessary in this domain to find landforms at different scales within elevation data. Finally, FloodTrace is the only system the authors found that integrates topological segmentation into a web-based application providing unprecedented accessibility and functionalities to crowdworkers.

\begin{figure*}
    \centering
    \includegraphics[width=0.95\textwidth]{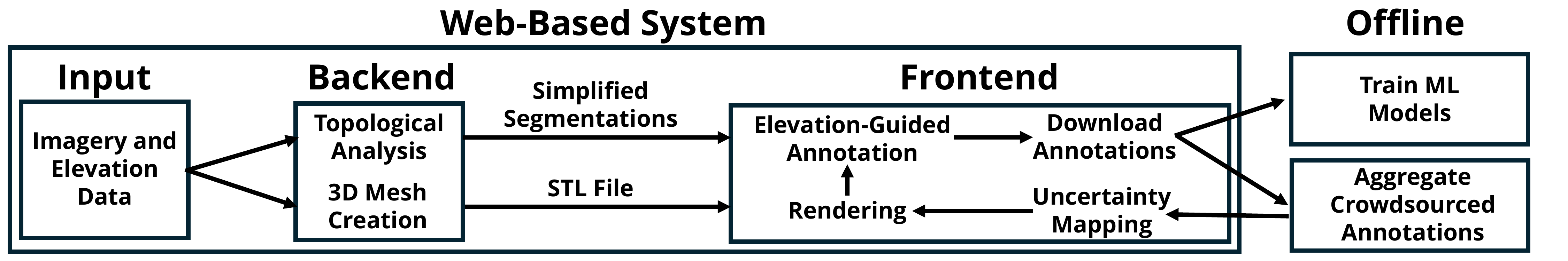}
    \vspace{-0.5em}
    \caption{\label{fig:pipeline_sketch}
    An illustration of the full processing pipeline for FloodTrace. \highlight{Input is given to the application in corresponding RGB imagery and elevation data. The elevation data is used in the server backend to create a 3D mesh which is combined with the RGB imagery as shown in }\autoref{fig:mesh_creation}\highlight{. The elevation data is processed by the backend to create simplified topological segmentations at different levels of detail, as shown in }\autoref{fig:segmentation_example}\highlight{. The user then utilizes the elevation-guided tools described by }\autoref{sec:tools}\highlight{ on the rendering in the web frontend. These annotations can then be downloaded for training machine learning models directly in the case of expert-labeled data, or for aggregation in the case of crowdsourcing. Aggregated annotations can be given as input to our application to be used with our uncertainty visualization tool for review and correction, as shown in }\autoref{fig:aggregate}\highlight{, before being used to train machine learning models.}
    }
    \vspace{-1.75em}
\end{figure*}

\section{Design}
\label{sec:design}
Before creating our solution, we interviewed domain scientist collaborators (i.e., machine learning researchers building models for flood extent mapping) to understand the state-of-the-art and requirements for this project. \highlight{Our collaborators were two junior researchers (PhD students) and one senior researcher (associate professor) who had each been working in the domain for multiple years. In the initial interview, we simply allowed these collaborators to voice their goals for an annotation tool without providing guidance, to not lead their conclusions. In this discussion, they highlighted the need for elevation-guided annotation, then focused on their usage of the state-of-the-art}~\cite{adhikari_elevation_2022}\highlight{ and its limitations. From this interview, we created requirements R1 and R2. From then, we began developing our application while continuously integrating feedback, as advised by best practices}~\cite{sedlmair2012design}\highlight{. To accomplish this, we conducted several unstructured interviews via in-progress meetings to demo the tool and request feedback. Towards the end of the development cycle, the application was also shared with the collaborators via a web deployment for them to experiment further. While much of the feedback received led to usability improvements, early progress meetings also led us to the creation of requirement R3, as the researchers emphasized the labor hours required for annotation.} We list the complete requirements we developed for building an effective application for flood annotation for ML training data:
\begin{itemize}
    \item R1: Image annotation decisions must be informed by elevation data. Elevation awareness enables robust annotation of regions that are ambiguous with only imagery data and more accurate annotation along elevation borders, naturally leading to higher quality training data for ML models that rely on elevation data. For accuracy and productivity, there should be intuitive and interactive visualization of both elevation data and flooded imagery, which is missing from previous work\cite{adhikari_elevation_2022}. 
    \item R2: The annotation application must fully support the large data sizes of study areas. Previous annotation work\cite{adhikari_elevation_2022} can only process small patches at a time, lowering productivity and leading to potential inaccuracies on patch borders. 
    \item R3: Annotation tools must be efficient and quick. Annotating a single flooded region usually involves labeling millions of pixels and multiple hours of work, and study areas commonly consist of many regions. Because of this, the acquisition of ground truth labels for training is extremely time-consuming. Tools that increase annotation productivity will have a large impact on improving ML researchers' workflows.
\end{itemize}

Orthogonally, we also discussed our collaborators' interest and concerns with crowdsourcing annotations to be used as training data. They expressed confidence that, while working on flood forecasting requires extensive experience and domain knowledge, data for flood detection (i.e. flood extent mapping) can accurately be labeled without expert knowledge by using aggregation of annotations from many participants. This is supported by previous work\cite{liang_flood_2018} which shows that aggregate annotation data seems to converge to the highest quality around 20-25 annotators for a particular study area, although this number likely changes depending on the difficulty of annotation region, background of participants, and tools being used. While our collaborators were optimistic about crowdsourcing, after discussion we were left with another requirement for our annotation system to support crowdsourcing properly:
\begin{itemize}
    \item R4: Aggregate annotation data must be presented in an understandable way for researchers, and provide the ability to correct inaccuracies. When dependent ML models make mistakes or do not behave as intended, it is important for researchers to be able to easily check model training data for errors and remedy them, especially when relying on crowdsourced data. Our application should provide effective visualization of aggregate annotations along with tools to improve them quickly. 
\end{itemize}

We implement our system to address these requirements in \autoref{sec:method}.
\begin{figure}[t]
\centering
    \includegraphics[width=\columnwidth]{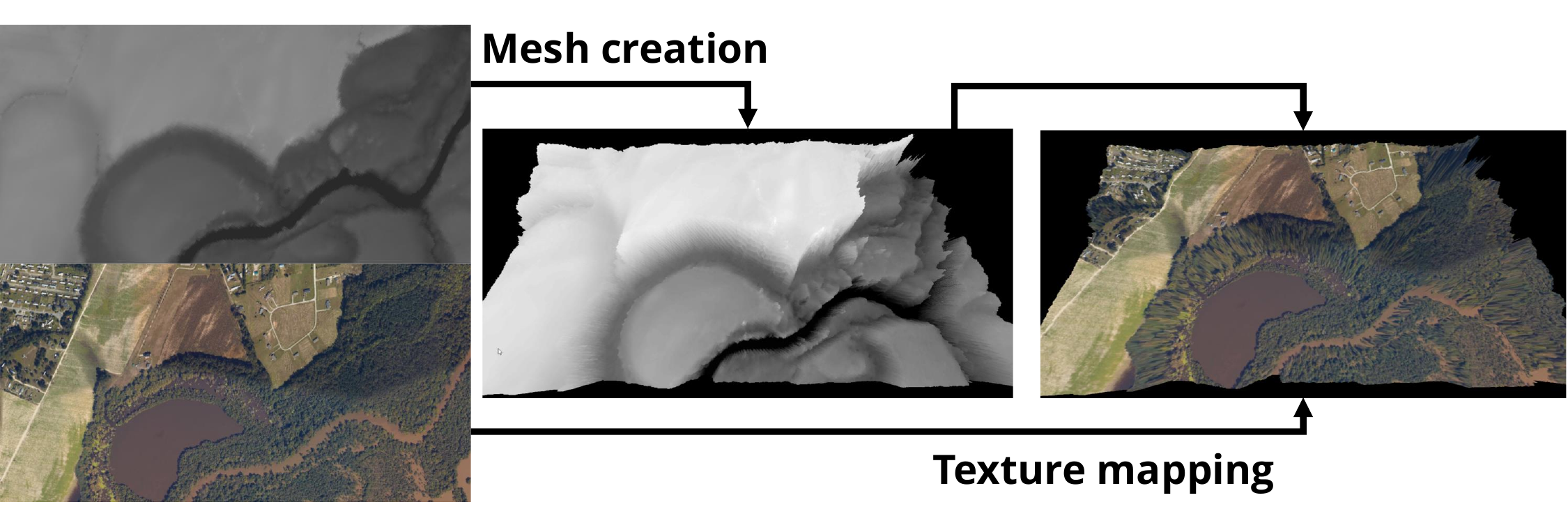}
    \caption{\label{fig:mesh_creation}
    An example showing mesh creation and rendering with aerial imagery RGB texture on a 1000 $\times$ 500 resolution example.}
    \vspace{-1.75em}
\end{figure}
\section{Implementation}
\label{sec:method}
FloodTrace consists of a web-based system that addresses the requirements gathered in \autoref{sec:design} for both crowdsourcing and direct annotation by researchers. To address R1, our system provides interactive 3D visualization of aerial imagery and DEM data (\autoref{sec:rendering}). R1 and R3 are both addressed by providing efficient elevation-guided semi-automatic annotation tools (\autoref{sec:tools}). R4 is addressed with aggregate annotation visualization for viewing uncertainty and making corrections (\autoref{sec:uncertainty}). For R2, our application is built using the GPU-accelerated WebGL\cite{WebGL} backend in Three.js\cite{threejs}. In our experimentation, the largest data size we have tested is 12000 $\times$ 12000 pixels, which is much larger than flood extent mapping regions. 
We give a brief overview of our UI design in \autoref{sec:user_interaction}.

A complete pipeline overview is given in \autoref{fig:pipeline_sketch}. Researchers first provide corresponding aerial imagery and elevation data to the server backend for computing the 3D mesh and topological data structures necessary for our methods. For crowdsourcing, these can easily be precomputed and served directly at a deployed frontend site, as in our user study, removing the need for crowdworkers to be given raw data or access to the backend server. Once data is served to the frontend, annotations can be made using elevation-guided tools on an interactive 3D rendering. Once the annotation process is complete, those can be downloaded and used directly to train ML models. In the case of crowdsourcing, annotations can be downloaded and submitted to researchers for aggregation. Researchers can then use these aggregated annotations with uncertainty visualization tools in the frontend to review and improve them before using them to train ML models.


\subsection{Rendering}
\label{sec:rendering}
An important feature to address R1 is providing the user with an interactive 3D visualization to inform their annotation. Upon receiving the input of RGB imagery data and DEM data, our application creates a 3D mesh in the backend as shown in \autoref{fig:pipeline_sketch}. We utilize the HMM heightmap meshing utility \cite{hmm} to triangulate a mesh from a given elevation image using the method of Garland and Heckbert \cite{garland1995fast}. The generated STL file is sent back to the web application, where it is visualized with the RGB imagery data texture mapped onto the mesh, as shown in \autoref{fig:mesh_creation}.  

We acknowledge that it would be possible to render the height field directly \cite{kerbltriangle2022, dupuytree2020, cornelheightfield2023}, however in our initial testing, we found direct height field rendering in Three.js to impact the interactivity of our system for large datasets. Pre-computing a triangulated mesh allows for easy, interactive visualization on our front end. While triangulating a surface from DEM data has been shown to introduce more error than higher-order methods \cite{kidnerinterpolation2003} and recent work has been able to reconstruct volumes for flooded terrains with higher accuracy \cite{cornelflood2019, boorboorstorm2024}, we find this loss acceptable for our use case as the mesh itself is only used for exploration. Our elevation-guided annotation tools use the full-resolution underlying elevation data. 

We initially experimented with fly-through camera controls for our rendering but decided against them after negative feedback from our collaborators. Instead, we implement orbit controls, modeling other applications that handle 3D data such as Paraview. This camera control method focuses on the annotation mesh, providing a more intuitive experience for interacting with the 3D visualization. The mouse scroll wheel controls the zoom, left click and drag rotates the camera around the mesh, and right click and drag pans the camera. In order to accommodate users working on laptops with only a trackpad who cannot right-click and drag easily, we also give an option for double-clicking on the mesh to pan to the area clicked. The camera view can be reset to its starting position with a click of a button in the menu.


Our collaborators expressed that having the ability to view the RGB imagery on a flat 2D mesh can be beneficial in certain situations, so we incorporate an option that allow users to seamlessly switch between 3D and 2D views. As users annotate a region, the pixels labeled as flooded or dry are overlaid with a translucent color mask of red or blue respectively. These colors were chosen to match the convention in the domain \cite{adhikari_elevation_2022}, with bright red and blue uncommon in satellite imagery. This overlay can be toggled, and it is helpful to do so after annotating large features with semi-automatic tools to ensure the area has been annotated correctly. These mesh appearance functionalities are implemented with custom WebGL fragment shaders for our Three.JS rendering.


\begin{figure}[t]
    \centering
    \includegraphics[width=\columnwidth]{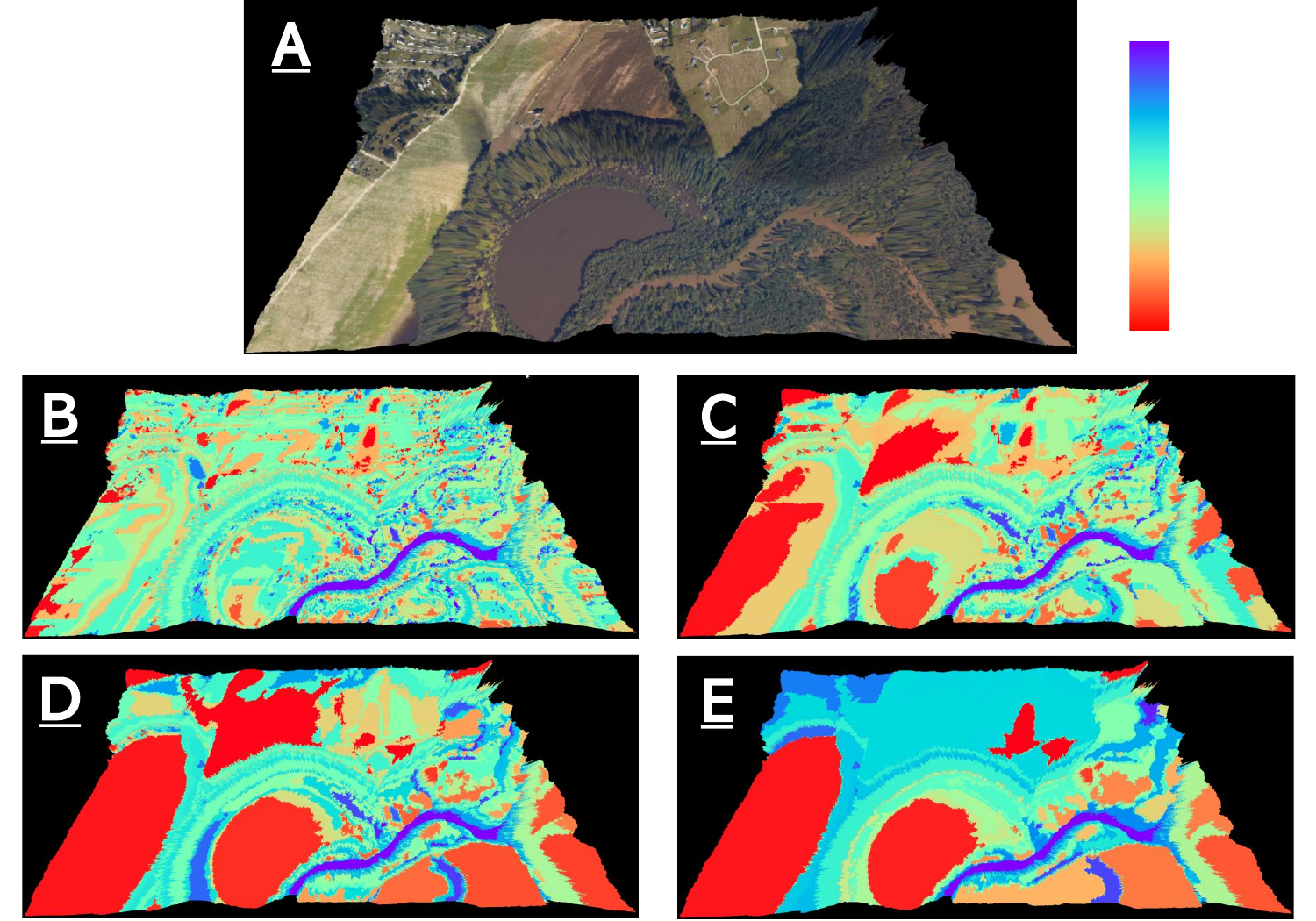}
    \caption{\label{fig:segmentation_example}
    An example region (A) with contour tree segmentations after no simplification (B) and at persistence thresholds $\epsilon$ = 0.02 (C), 0.04 (D), and 0.08 (E) of the elevation data function range. As seen in (B), segmentations produced without simplification are too noisy to be useful, while those in (E) correspond to data features such as the hills, lake, and river. Segmentations colored by rainbow colormap. \highlight{Our application allows users to label these segmentations in one click to quickly annotate features as flooded or dry.} }
    \vspace{-1.75em}
\end{figure}

\subsection{Elevation-Guided Annotation Tools}
\label{sec:tools}
Following R1 and R3, FloodTrace utilizes elevation data for annotation tools that are able to quickly create accurate, robust labels for large flooded regions. These tools are built for intuitive use by both domain experts and crowdworkers. We provide both extension-type and segmentation-type semi-automatic methods in the form of the BFS (\autoref{sec:BFS}) and topological segmentation (\autoref{sec:segmentation}) tools respectively. 

\subsubsection{BFS Tool}
\label{sec:BFS}
This method of extension-type semi-automatic annotation was first proposed by Adhikari et. al \cite{adhikari_elevation_2022}, and we adapted it as a state-of-the-art tool in our application. This tool is used by annotating a seed pixel, which is then extended in all directions by connected downstream or upstream pixels (depending on whether the label is flooded or dry) in the elevation data. This accurately labels flooded or dry regions by taking advantage of the physical constraint that if a location is flooded, then its adjacent locations with a lower elevation will also most likely be flooded. In the same way, adjacent locations of dry areas that have higher elevations will most likely be dry. This approach accurately annotates regions along elevation border pixels, as the BFS stops where the elevation becomes higher or lower than the connected pixel, depending on if flooded or dry is being labeled. This approach also addresses the difficulty of annotating ambiguous pixels in the RGB imagery, such as those covered by tree canopy or clouds, by automatically labeling them using neighboring pixels.  

Thanks to the 3D mesh rendering in FloodTrace, the user can purposefully select high-elevation flooded pixels or low-elevation dry pixels in order to label as large of a downstream or upstream area as possible, improving efficiency. Our application also extends the method with a polygon BFS tool, where one can select points to form an arbitrary polygon, fill the polygon, and run a BFS selection from all points on the borders. This feature allows larger areas to be selected quickly while ensuring that the annotation's border pixels correctly correspond to elevation changes.

\subsubsection{Topology Segmentation Tool}
\label{sec:segmentation}
We utilize topological data analysis to create a segmentation-type tool that can quickly annotate regions according to multi-scale features in the elevation data. To do this, FloodTrace utilizes the Python binding of the Topology Toolkit (TTK) \cite{tierny_ttk_2018}, specifically its functions for simplification by persistence (which implement the works of Tierny and Pascucci\cite{tierny_simplification_2012} and Lukasczyk et al.\cite{lukasczyk_simplification_2020}) and contour tree creation (which implement the work of Gueunet et al. \cite{gueunet_tree_2017, gueunet_tree_2019}). Given input data, the backend first uses persistent homology to compute simplifications of the data at varying thresholds, then creates a set of contour tree segmentations from the simplified data. These segmentations are sent to the frontend, where they can then be used for annotation. We show an example region with contour tree segmentations from data simplified at different levels in \autoref{fig:segmentation_example}. Notice how the simplified segmentations (such as E) capture useful terrain features for flooded and dry areas, such as the large hill at the bottom left, lake and river in the middle, and pond in the bottom right. 

By providing the segmentation set for selection, users can quickly annotate large, obviously flooded and dry features, while maintaining the detail necessary to accurately label smaller features, allowing for robust annotation of features at different levels of detail. Annotation is done simply and quickly with the segmentation tool in our application by applying flooded or dry labels to the segment currently under the cursor. The user can quickly switch between different simplification levels in the UI, interactively changing the segmentations that will be used for selection to that level. Users must switch between simplification levels in order to ensure that they do not label an area so large that it includes both flooded and dry regions or so little that they do not fully capture a flooded or dry feature, so using this tool does take some experimentation. Example annotation using this tool can be seen in our supplementary video.


Based on demo feedback from our collaborators, we provide two options for visualizing the segmentations of a dataset. The first option is to paint the borders of the current segmentations on the mesh in white, which effectively shows contours of the data simplified at the current level. This is used to quickly see which simplification level leads to the desired level of detail by viewing how segmentation borders change when cycling through the levels. The second option is to highlight in red or blue (depending on flooded or dry potential label) the segment currently hovered over by pressing the highlight key. This is used to quickly check the coverage of the selected segment to confirm it contains fully flooded or dry imagery before labeling.



To keep the interaction simple for the user study, we provided six simplification levels based on thresholds at logarithmic steps along the datasets' normalized function range (i.e. $\epsilon$ = 0, 0.01, 0.02, 0.04, 0.08, 0.16). While we found these sufficient for all of our study regions, necessary simplification could vary based on a particular researcher's dataset. To support this, users can choose the number of thresholds and persistence values themselves when submitting input data, as well as query the backend for more simplified data as needed.

\begin{figure}[t]
\centering
    \includegraphics[width=\columnwidth]{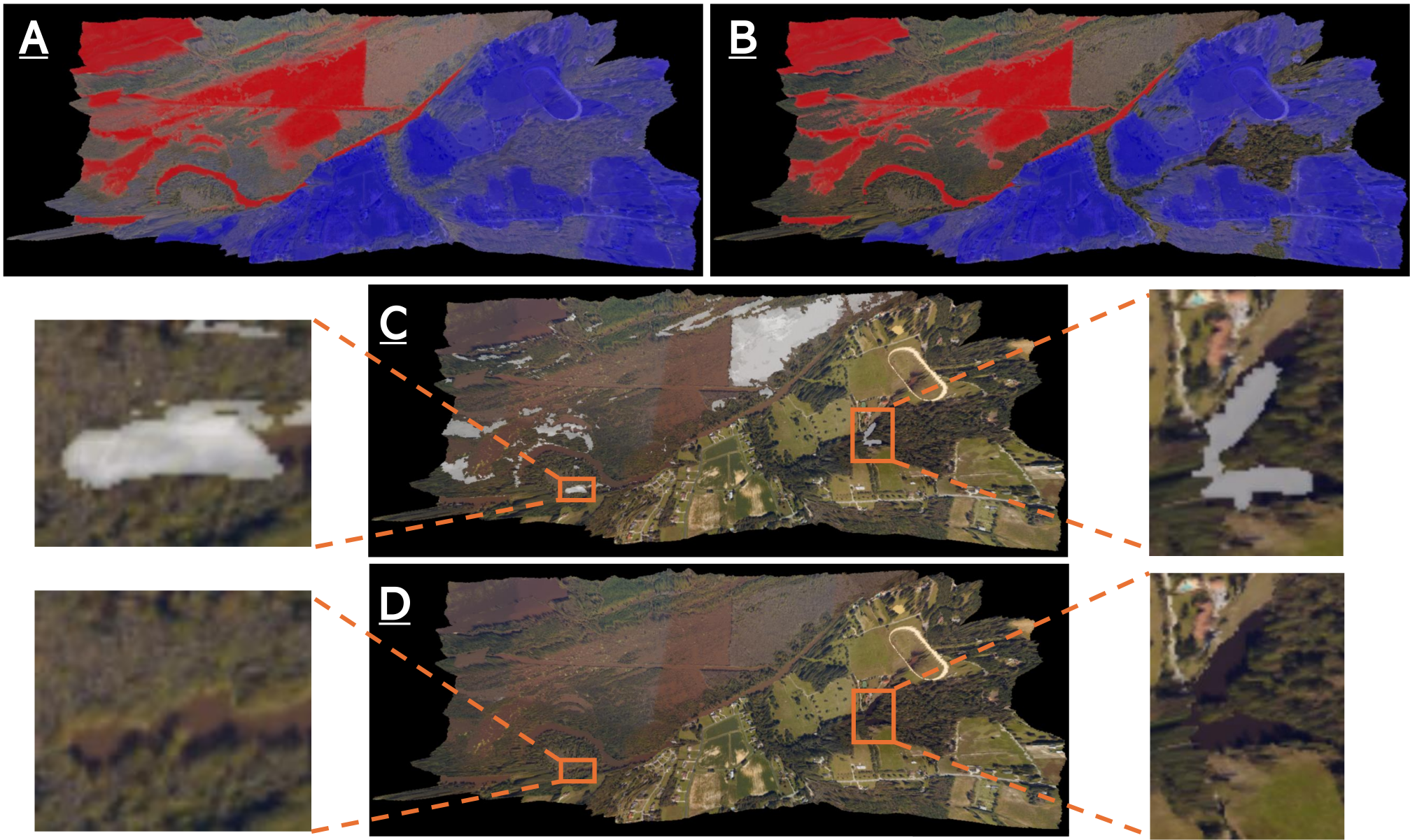}
    \caption{ \label{fig:aggregate} Visualization of aggregated crowdsourced data from 45 participants. The aggregate view is shown in A) with no certainty threshold and in B) with a certainty threshold of 0.6. This view can quickly show the areas where the group of annotators were confident in labeling flooded or dry. The variance view is shown in C) with a threshold of 0.7 and compared in D) to a view with no annotation texture. With this view, it is easy to identify regions where there was high annotator disagreement, such as those shown in the orange borders. These selected regions are obviously flooded, and so the researcher can quickly correct these areas by labeling them. We explore how this can improve model performance in \autoref{sec:evaluation}.}
    \vspace{-1.75em}
\end{figure}

\subsection{Aggregated Annotation Visualization}
\label{sec:uncertainty}
In order to address R4, FloodTrace offers the capability for researchers to visualize aggregated crowdsourced annotations and fix inaccuracies before training dependent models. We accomplish this through two novel views that are inspired by work on uncertainty visualization in mapping applications. In prior work\cite{hengl_uncertain_2003, hengl_uncertain_2006, sanyal_uncertain_2009}, the uncertainty of 2-dimensional variables was communicated effectively through inverse mapping of color saturation to the magnitude of uncertainty; when the value of the variable is more uncertain, the pixel's color is made to have lower saturation. We follow this standard, with uncertainty for pixels in our aggregate visualization defined by scarcity of labels or variance between annotators' labels in the given set of annotations. 

Before visualization, each annotation in a given set is transformed into a 2-dimensional array where red (flooded) pixels are made -1, transparent (unlabeled) pixels are made 0, and blue (dry) pixels are made 1. For our first view, the aggregate view, we find the mean for each pixel across annotations, giving a value in [-1, 1] where a score of -1 means all annotators agreed the pixel is flooded, 1 means all annotators agreed the pixel is dry, and scores closer to 0 mean the annotators either disagreed or chose not to label the pixel. These values are then visualized as a color-mapped texture on the 3D rendering within our application. We follow the standard of inversely mapping uncertainty to saturation in our colormap, but because the underlying RGB imagery also contains important information for the researcher, we extend this by additionally mapping uncertainty to texture transparency. Pixels that annotators uniformly labeled as flooded or dry are colored with a more opaque red or blue, while this color becomes more white and transparent the more uncertain annotators were about its label. 

To simplify the texture being mapped, a sliding tool in our interaction menu gives users the ability to threshold pixels by their certainty. This collapses pixels; values to 0 if their absolute value is not greater than the threshold, effectively hiding the values since 0 is fully transparent. An example of this view aggregated from 45 unique annotations from our user study is shown in \autoref{fig:aggregate} (A, B). This view was developed with collaborator feedback as a way to quickly understand the labels being used for training data and identify possible sources of error, helping to address R4. In addition, while dependent models can be trained using scored soft labels of flood and dry scores, the ability to explore certainty thresholds in our view allows researchers to find values for binarizing aggregate data into hard labels.

In order to fully address R4, we create another view specifically for revising aggregated crowdsourced annotations. This view, the variance view, highlights the areas where annotators disagreed the most, as these are likely to need revision. It does this by displaying the computed variance for the set of annotations. We use a similar color mapping as before, with higher uncertainty (in this case variance) corresponding to white color. Using a white color for areas of interest can clash with the satellite imagery when it contains white features such as buildings, so we also provide the option to use a low saturation pink. Because we want to focus the user on the areas of high uncertainty for correction, we inversely map uncertainty to texture transparency. This view is shown in \autoref{fig:aggregate} (C). With this view, researchers can easily check the areas of high variance and annotate them with our brush, BFS, or segmentation tools just as they would for a normal annotation task. The crowdsourced annotations can then be downloaded with hard correction labels applied to these annotated areas to improve model training. 

\begin{figure}[t]
\centering
    \includegraphics[width=\columnwidth]{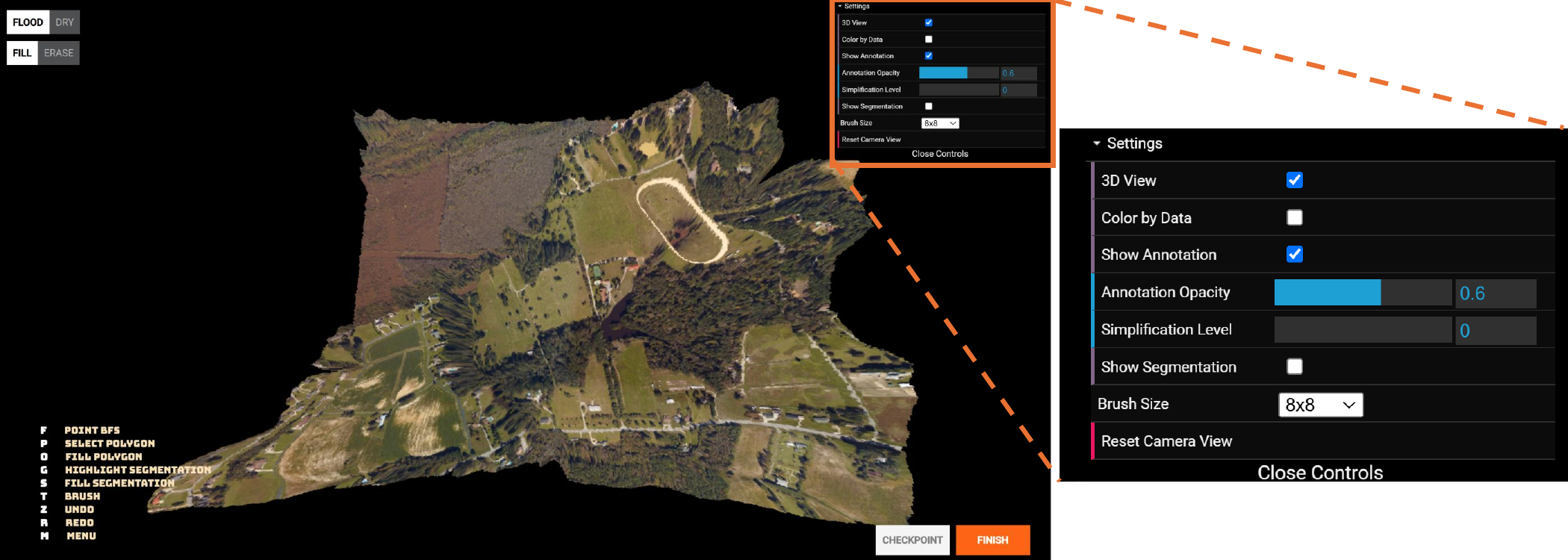}
    \caption{ \label{fig:ui} UI for our application. Settings blown up for readability.}
    \vspace{-1.75em}
\end{figure}
\subsection{User Interaction}
\label{sec:user_interaction}
The UI for FloodTrace is shown in \autoref{fig:ui}, with an interaction menu in the top right, toggles between erase/fill and flood/dry selection in the top left, key reminders in the bottom left, and checkpoint/finish buttons in the bottom right. The on-screen UI is limited in order to simplify the user experience and allow the annotation mesh to take up the full view.



Users make annotations by aiming the mouse cursor at an area of the mesh and pressing the chosen key of any of the annotation tools. Along with the semi-automatic tools, we also provide a brush to manually paint and erase annotation. Point BFS, segmentation, and brush tools immediately annotate the area underneath the cursor using their respective algorithms on key press. Polygon selection with BFS occurs by pressing a key to repeatedly select desired points on the polygon to be drawn and then pressing another key to confirm the selection. This then fills the selected polygon and runs the BFS method on the pixels of its boundaries. 

The current annotation type/color is toggled with a switch on the top left of the UI, which affects whether the BFS method selects upstream or downstream pixels. Users can also toggle annotation mode from 'Fill' to 'Erase' in order to use all the different annotation tools to remove annotation color rather than paint on the texture. Undo and redo functionality are given to allow users to take back accidental or ill-thought-out annotations, and put them back again. 

As users utilize the various tools in FloodTrace, all of their actions are logged with enough detail to replay their session completely given an output log. This gives the ability to create checkpoints during annotation, where the state of the current session is downloaded as a JSON log file and can be uploaded in the future to restart the session at that state. Sessions can also be started with a given annotation texture. When a user selects to finish their session, their log file, annotation texture, and metadata about the session are downloaded.
\section{Evaluation}
\label{sec:evaluation}
In this evaluation, we study FloodTrace through an experimental user study explained in \autoref{sec:setup}. We use the results of this study to assess the efficiency of our topological segmentation tool and the value of our aggregate visualization and correction strategy. To do this, we first analyze the accuracy and annotation speed of participants (\autoref{sec:user_study}). We then use participant annotations as training data for flood detection models and assess model quality (\autoref{sec:machine_learning}).

\subsection{User Study Setup}
\label{sec:setup}
The dataset used for the study consists of flooded regions of North Carolina during Hurricane Matthew in 2016, with high-resolution aerial imagery from the National Oceanic and Atmospheric Administration National Geodetic Survey \cite{noaa} and corresponding digital elevation model (DEM) data from the University of North Carolina Libraries \cite{dem}. All data was resampled to a resolution of 2 square meters per pixel, which is common in this domain \cite{jiang_markov_2019, sainju_markov_2022, xie_markov_2018}. Eight separate regions were chosen with dimensions between 4104 $\times$ 1856 pixels and 6472 $\times$ 3136 pixels, and these regions were split into quadrants for smaller workloads for participants of the study.

297 graduate students of a machine learning course volunteered to participate in this study, of which 266 followed through with final submissions. Before beginning annotation work, participants watched an instructional video explaining how to use all of the features of FloodTrace and how to determine whether to mark areas as dry or flooded. They were then given access to a web application which they could visit at any time within the following two weeks to complete the work assigned to them. On the application, all interactions made by the students were logged, and these logs were submitted with output annotations on completion. Each student was asked to complete five annotation tasks, with each task requiring annotating at least $60\%$ of a quadrant of one of our chosen regions as flooded or dry, leaving any remaining pixels as unlabeled. These tasks were created for two specific experiments, in addition to providing a reasonable number of annotations and interaction logs to draw insight from. 

The goal of the first experiment was to fairly compare the effectiveness of the annotation tools, ensuring that participants became familiar with both semi-automatic tools. This experiment consisted of three annotation tasks performed on the same quadrant. For the first, the user would perform annotation while restricted to only one of the semi-automatic tools. Next, they would complete the same annotation task while restricted to the semi-automatic tool they did not have in the first task. Last, they would complete the annotation task again with access to both semi-automatic tools. Participants were divided evenly so that half started with the BFS tool and half started with the segmentation tool. Brush tools were provided in all three tasks to allow students to make fine-grained corrections or label features that were difficult to annotate semi-automatically.

The second experiment was meant to test the effectiveness of the 3D mesh in improving participants' insight and accuracy. This required users to perform two annotation tasks on the same quadrant, although different form the one used in the first experiment. In this experiment, students would annotate while restricted to only a 2D view of the aerial imagery, then they would annotate the same quadrant with the 3D mesh and full 3D view interactivity. Both tasks for this experiment had full access to brush and semi-automatic tools. 

At the conclusion of this study, we received 266 submissions, with 259 of those including all five tasks that the participant was assigned. Aggregating all submissions, each quadrant in our dataset was annotated on average 41 times, 1,321 total annotations were collected, and participants collectively annotated over 3.5 billion pixels. We open source this dataset of annotations as a contribution of this paper, along with metadata for each annotation. 

\subsection{User Study Analysis}
\label{sec:user_study} 
\highlight{In this section, we analyze the results of our user study to draw conclusions about how the usage of different annotation tools (brush, BFS, and segmentation) affected annotation speed and accuracy. Specifically, we test the hypotheses that elevation-guided tools increase accuracy, while the segmentation tool decreases the time needed for annotation. We also assessed how providing users with a 3D rendering affected their accuracy, using the two annotation tasks from the second experiment in }\autoref{sec:setup}\highlight{, but found no statistically significant results so omit these results for space. In }\autoref{sec:analysis_setup}\highlight{ we first define our metrics and describe how the data was preprocessed. 
Next, to compare the tools against each other and quantify the differences in speed and accuracy, we divide annotations into groups depending on which tool was used for the majority of labeling and compute differences between groups in }\autoref{sec:tool_analysis}\highlight{ . For all statistical tests, we select a 95\% confidence interval for significance, then use the Benjamini-Hochberg (BH) procedure to correct for multiple testing by limiting the false discovery rate to 0.05 in each family of tests. Necessary p-value thresholds for significance computed by BH (}$\alpha$\highlight{) are presented with each table.}
\subsubsection{Tool Analysis Setup}
\label{sec:analysis_setup}
Because the goals of R1 and R3 are to increase annotation speed and accuracy, these are the metrics we study. We measure annotation speed simply as the time taken to complete an annotation. We measure annotation accuracy by comparing participant annotations against reference annotations that were manually labeled either by our domain expert collaborators, or members of our team after being trained. To avoid penalizing unlabeled pixels, we only consider pixels that were labeled as flooded or dry in both the participant and the reference annotation when measuring accuracy. The formula we use to compute accuracy is shown below. TF and TD denote true flooded and true dry (pixels with the same label in participant and reference), and FF and FD are false flooded and false dry (pixels with opposite labels in participant and reference):
$$\frac{TF + TD}{TF + TD + FF + FD} \times 100\%$$
An accuracy percentage of 0\% here means all participant-labeled pixels disagreed with all corresponding labels in the reference, 100\% means all user-labeled pixels agreed with corresponding labels in the reference, and 50\% means half of the user-labeled pixels agreed with corresponding labels in the reference. Note that a score of 50\% can be achieved by labeling pixels randomly since there is a 50\% chance of guessing a flooded or dry label correctly. 


Before statistical analysis, we first conducted data cleaning to remove potential spoilers from the dataset, as we found some participants to misunderstand the labeling task or leave the application open without progress for extremely long periods. We computed the means and standard deviations for annotation completion time and accuracy, and used these to compute a threshold at which to discard annotations. We chose a standard z-score of 3 as our cutoff point (meaning 3 standard deviations from the mean, or falling outside of the 99.7\% confidence interval), and were left with thresholds of 17.6 hours for annotation time and 55.4\% for accuracy. Because the variance of annotation times was much higher than the variance of accuracies, the time threshold could likely be more aggressive; the median completion time was only 32 minutes, so it seems probable that times above a few hours would be the result of idling on the page. We investigated using z-scores of 1 and 2 to create stricter time thresholds (corresponding to 7.0 and 12.3 hours respectively), but found that these led to the same statistically significant conclusions as using a z-score of 3, only more pronounced, so we present here the results using the more inclusive 17.6 hour cutoff. With our time and accuracy thresholds, we discard a total of 40 annotations and leave 1,281 activity logs for analysis.

\subsubsection{Tool Analysis Results}
\label{sec:tool_analysis}
\highlight{To study the impact of different tools on our metrics, we use the participants' activity logs to group these annotations by which tool was used for a majority of pixels. We can then compare these groups' accuracies and annotation time by using Welch's t-test, which computes whether the differences in the metrics are significant for groups of unequal variances and sizes.} We create four groups based on tool usage as described below:
\begin{itemize}
    \vspace{-0.5em}
    \item[] \textbf{Brush}: Annotations where the brush tool was used to label $>50\%$ of all labeled pixels
    \vspace{-0.5em}
    \item[] \textbf{BFS}:  Annotations where the BFS tools (polygon and point selection) were used to label $>50\%$ of all labeled pixels
    \vspace{-0.5em}
    \item[] \textbf{Segmentation}: Annotations where the segmentation tool was used to label $>50\%$ of all labeled pixels
    \vspace{-0.5em}
    \item[] \textbf{Elevation Guided}: Annotations where elevation-guided tools (segmentation and both BFS selections) combined were used to label $>50\%$ of all labeled pixels. \highlight{This group is used to directly compare elevation-guided tools against the brush}
\end{itemize}
\begin{table}
  \centering
      \vspace{-1.5em}
    \relsize{-1}{
  \begin{tabular}{|c|c|c|c|c|}
    \hline
    & 
    \textbf{Brush}  & 
    \textbf{BFS}  & 
    \textbf{Segment.} &
    \textbf{Elevation Guided}
    \\ \hline
    \textbf{Mean}
    & 90.52 & 91.76 & 91.64 & 91.67 \\ \hline
    \textbf{P-Value vs Brush}
    & - & 0.092 & 0.031 & 0.020 \\ \hline
  \end{tabular}}
    \vspace{-1em}
    \caption{\label{tab:acc}%
    Mean accuracies and p-values from t-tests against the brush group. We find all three groups that rely on elevation-guided tools to be more accurate than the group using the manual tool. \highlight{The }$\alpha$\highlight{ needed for significance (computed by BH) for this family of tests was 0.033.} All comparisons are statistically significant except the BFS group, likely because the group had a much smaller sample size.}
    \vspace{-1.75em}
\end{table}
\begin{figure}
\vspace{-2em}
\centering
    \includegraphics[width=\columnwidth]{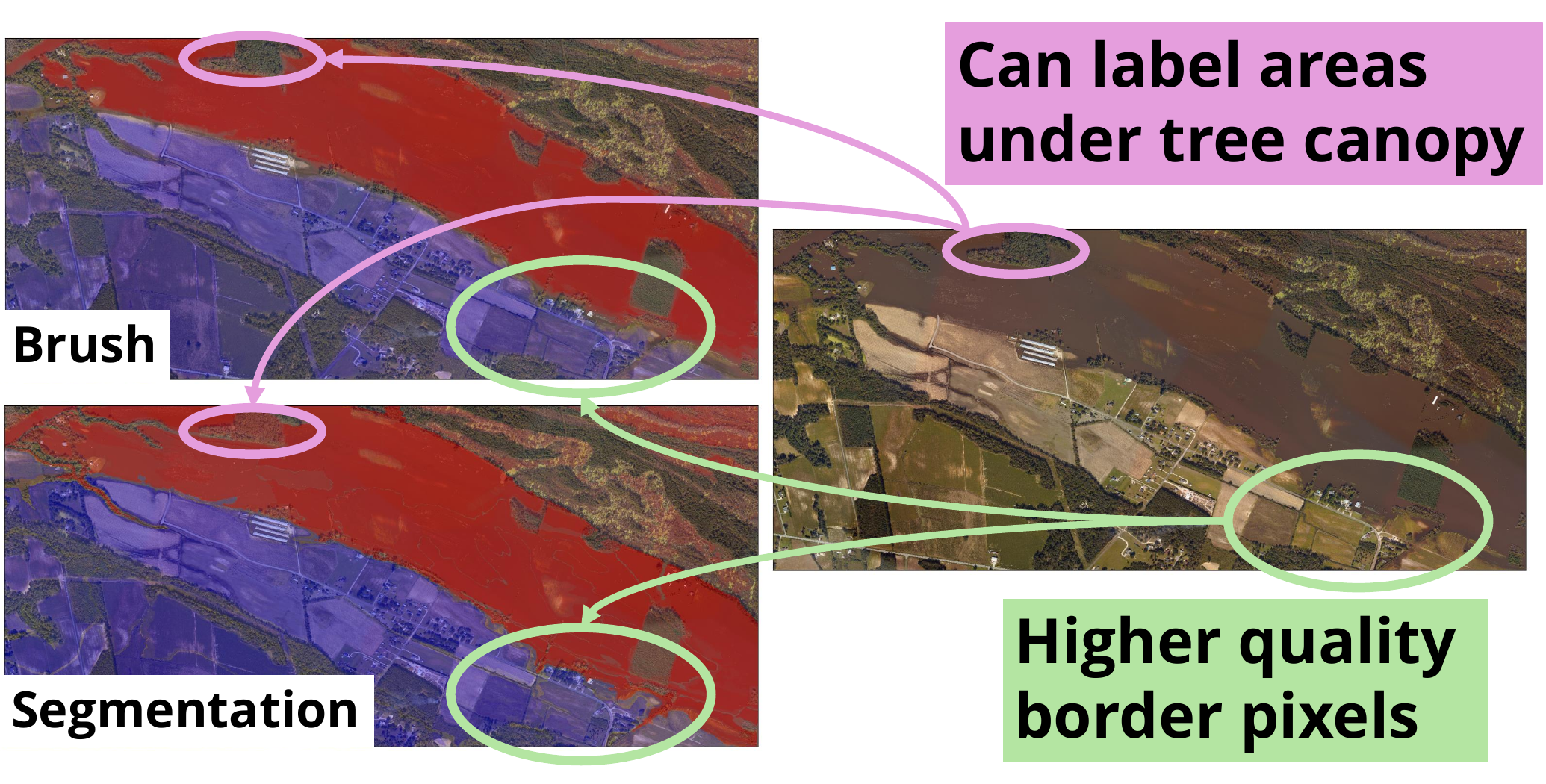}
    \vspace{-1.75em}
    \caption{\label{fig:aggregate_compare}
    Aggregated annotations from brush and segmentation groups on a study quadrant, visualized using the flattened aggregate view with a certainty threshold of 0.33. In green, notice how the annotations from the brush group do not label the border pixels between the flooded and dry areas, while the annotations from the segmentation group correctly assign labels to these. Also, in pink, notice how the brush group leaves the selected tree canopy area unlabeled, while the segmentation group labels it flooded.
    }
\end{figure}
In order, these groups ended up with sizes of 787, 134, 346, and 494 annotations. We find that, even though the instructional video emphasized that participants should rely mostly on the elevation-guided tools, users gravitated towards the more simple and familiar brush tool. This likely shows an example of algorithm aversion \cite{dietvorstaversion2014, dietvorstaversion2016}, where users are proven to prefer human decision-making rather than algorithmic tools even when presented with algorithms that outperform humans. We acknowledge that this could be fixed in future deployments by forcing some usage of the elevation-guided tools, but we leave more in-depth approaches to overcoming algorithm aversion as future work. 

We provide the mean accuracies for each group in \autoref{tab:acc}. Because we test whether elevation-guided tools increase accuracy, we also present the p-values for their differences against the brush group. From these results, we find that those who relied on elevation-guided tools were able to produce more accurate annotations than those who relied on the brush tool. This shows that elevation-guided tools are effective for improving accuracy (and thus training data quality), even for untrained crowdworkers. We additionally checked for differences between the BFS, segmentation, and elevation-guided groups and found no statistically significant results.

To gain insight into the significance of accuracy differences between groups and explain how elevation-guided tools improve accuracy, we present a qualitative analysis of aggregated annotations from the brush and segmentation groups in  \autoref{fig:aggregate_compare}. For the chosen quadrant, there were 25 annotations in the brush group and 18 in the segmentation group. This example shows how the segmentation tool improves participant annotations by labeling border pixels and tree canopy. This is beneficial for dependent ML models, as elevation border pixels are especially important for elevation-guided training. 
\begin{table}
  \centering
    \vspace{-1em}
    \relsize{-1}{
  \begin{tabular}{|c|c|c|c|c|}
    \hline
    & 
    \textbf{Brush}  & 
    \textbf{BFS}  & 
    \textbf{Segment.} &
    \textbf{Elevation Guided}
    \\ \hline
    \textbf{Mean}
    & 79.61 & 77.73 & 48.07 & 56.74 \\ \hline
    \textbf{P-Value vs Brush}
    & - & 0.434 & 0.000 & 0.000 \\ \hline
    \textbf{P-Value vs BFS}
    & 0.434 & - & 0.007 & 0.036 \\ \hline
    \textbf{P-Value vs Seg.}
    & 0.000 & 0.007 & - & 0.129 \\ \hline
  \end{tabular}}
    \vspace{-1em}
    \caption{\label{tab:acc}%
    \highlight{Mean time taken and p-values from t-tests between each group. The }$\alpha$\highlight{ needed for significance (computed by BH) for this family of tests was 0.025. All comparisons are statistically significant except brush against BFS, and elevation-guided against BFS and segmentation. These results show that the segmentation tool greatly decreases annotation time.}}
    \vspace{-2em}
\end{table}
\begin{figure}
\vspace{-1.75em}
\centering
    \includegraphics[width=\columnwidth]{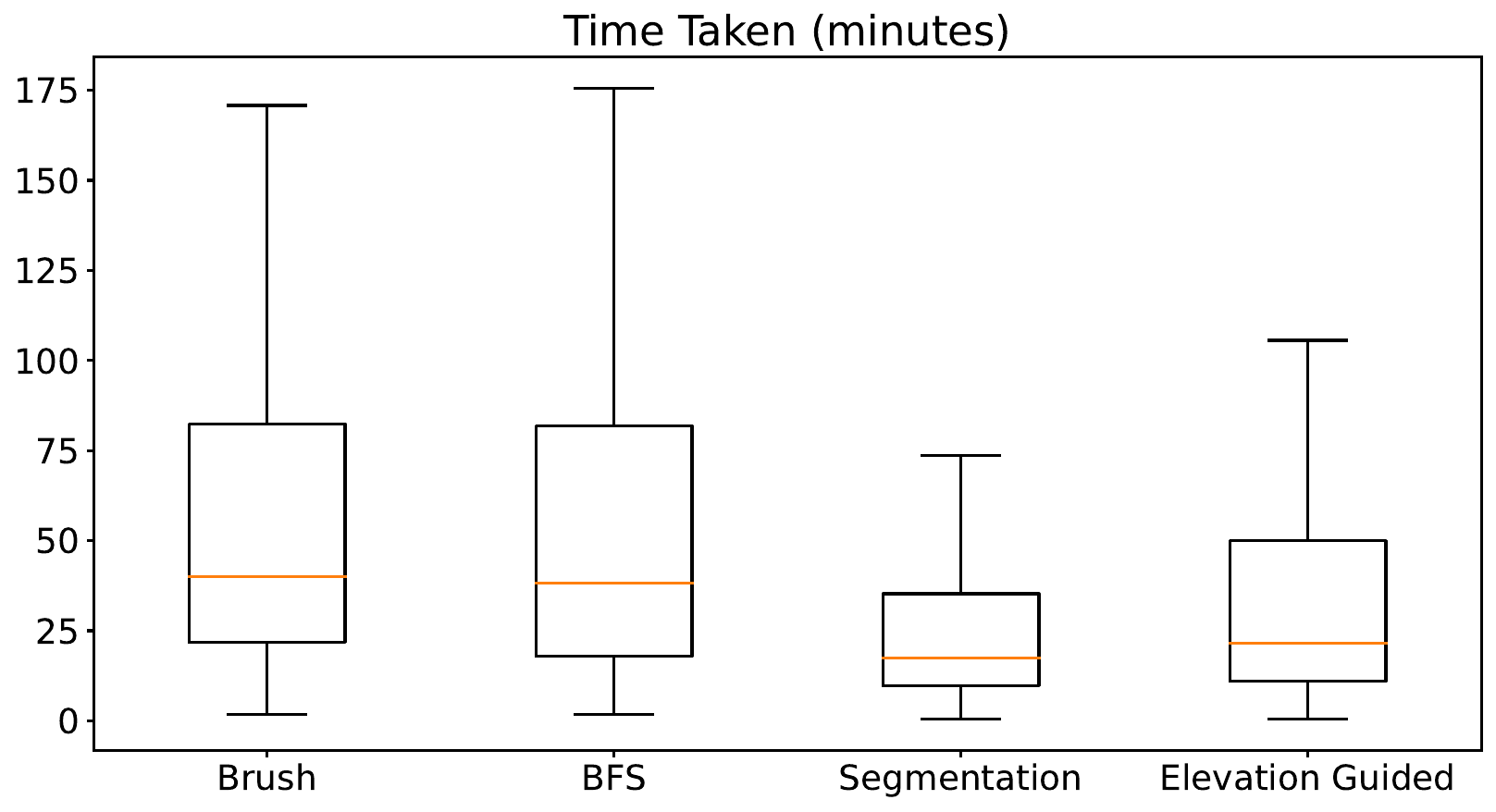}
    \vspace{-1.75em}
    \caption{\label{fig:time_taken}
    Box and whisker plot showing the time taken for annotations in the brush, BFS, segmentation, and elevation-guided groups. We find use of the segmentation tool drastically reduces the time taken to complete an annotation, with brush and BFS groups having median times of 39.97 and 38.18 minutes while segmentation and elevation-guided groups have median times of only 17.45 and 21.65 minutes.}
    \vspace{-1.75em}
\end{figure}

To test annotation speed of the tools, we provide the mean time taken for each group as well as the p-value for their differences against each other in \autoref{tab:acc}. From this table, we find that the segmentation tool was significantly faster than the other two tools. To better quantify the tools' differences, we also present annotation speed results in a box and whisker plot in \autoref{fig:time_taken}. This plot shows that annotations relying on the segmentation tool were much faster than those that did not, with the median time taken of the segmentation group being $2.29\times$ and $2.19\times$ smaller than the medians of the brush and BFS groups respectively. These results clearly support our segmentation tool as a method of quick annotation while preserving accuracy, and show it outperforms the BFS tool in efficiency. Importantly, these results show that even untrained crowdworkers can use this tool effectively. 

\subsection{Crowdsourced Annotations for ML Model Training}
\label{sec:machine_learning}
In this section, we show how FloodTrace can enable researchers to use crowdsourced annotations as high-quality training data through our aggregation and correction process, which we detail in \autoref{sec:example_aggregation}. We propose this step to replace the extremely time-consuming annotation step of researchers' workflows. To assess the training data quality of the results from aggregation and correction, we use study participants' annotations as training data for ML models and compare them against models trained with reference annotations \autoref{sec:prediction}.

\subsubsection{Example Aggregation and Correction}
\label{sec:example_aggregation}
We conducted an example of our aggregation and correction process on two randomly selected regions using all the annotations for those regions from our user study. We purposefully did not use any accuracy, time, or tool usage cutoffs when creating these sets of annotations in order to ensure there would be inaccuracies in the aggregate data. For each region, there was an average of 45 annotations per pixel. We input these sets to the application frontend, and one of our trained users utilized our aggregate and uncertainty views to correct each region. On average, $4\%$ of each region's pixels were corrected. This process was timed, and it took an average of 24 minutes to make each region's correction annotation. We compare this to the average time it took our trained users to create reference annotations using our application, which was 116 minutes per region. As training robust flood detection models requires many different annotated regions, this shows how the crowdsourcing and correction workflow can save researchers huge amounts of time compared to the typical annotation process. 

\subsubsection{Prediction Setup and Results}
\label{sec:prediction}
In order to use our crowdsourced annotations (corrected and uncorrected) to train our ML models, we pre-process them by turning each group of annotations into a training set of soft labels. This is done by computing the flood and dry scores for each pixel, where the flood score is the sum of the number of times it was annotated as flooded divided by the number of times it was annotated as either flooded or dry, with dry scores using the same formula. We ignore unlabeled pixels in each annotation to compute these scores. We experimented with binarizing training sets by giving hard labels of flood and dry to pixels with scores above a certainty threshold, but found better performance by training using these soft labels. This is likely because soft labels with scored values are able to encode the probability of pixels being flood or dry, which allows the model to learn more effectively.

For our deep learning models, we follow the architecture described by Adhikari et al. \cite{adhikari_elevation_2022} to create elevation-aware U-Net classification models to predict flooded and dry labels. Resulting predictions from this architecture take the form of a probability map, where pixels in the input are predicted with both flooded and dry probability values. We transform this into a binary class format for evaluation metrics by choosing the higher of the two scores for each pixel and applying that label. This gives us a fully labeled output, which we compare with our reference annotations to compute precision, recall, and f-scores for each class, along with accuracy as computed before. As the reference annotations include some unlabeled pixels, we ignore model predictions for these when computing metrics.

\begin{table}
  \centering
    \relsize{-2}{
  \begin{tabular}{|p{1.25cm}|c|c|c|c|c|c|}
    \hline
    \textbf{Group} & 
    \textbf{Class}  & 
    \textbf{Precision}  & 
    \textbf{Recall} &
    \textbf{F} &
    \textbf{Avg. F} &
    \textbf{Accuracy}
    \\ \hline
    \multirow{2}{*}{Uncorrected}
    & Dry & 0.902 & 0.959 & 0.930 & \multirow{2}{*}{0.895} &\multirow{2}{*}{90.6\%}\\ 
    & Flood & 0.916 & 0.810 & 0.859 &&\\ \hline
    \multirow{2}{*}{Corrected}
    & Dry & 0.973 & 0.947 & 0.960 & \multirow{2}{*}{0.934} &\multirow{2}{*}{94.4\%}\\ 
    & Flood & 0.881 & 0.937 & 0.908 &&\\ \hline
    \multirow{2}{*}{Reference}
    & Dry & 0.963 & 0.952 & 0.957 &\multirow{2}{*}{0.933} &\multirow{2}{*}{94.2\%} \\ 
    & Flood & 0.897 & 0.919 & 0.908 &&\\ \hline
  \end{tabular}}
    \vspace{-0.5em}
    \caption{\label{tab:f_scores}%
    Performance metrics for models trained using uncorrected, corrected, and reference labels as ground truths. \highlight{Results show that corrected crowdsourced labels lead to around equal accuracy and f-score as fully expert-created labels, while requiring much less work on the part of the researcher.}}
    \vspace{-2em}
\end{table}

We initialized three models, using the architecture described, and trained each of them in the same two regions. Each model used a different set of labels as its respective ground truths during training, with one model using the uncorrected crowdsourced labels, one using the corrected crowdsourced labels, and one using our reference labels. We then use these three trained models to predict flooded and dry labels for an unseen test region and compute our quality metrics. We show these results for each group in \autoref{tab:f_scores}. We find that, with the corrected crowdsourced labels, we are able to achieve around equal results (measured in accuracy and f-score) to our expert-labeled references and improve dramatically over the uncorrected labels. 

\section{Conclusion and Limitations}
\label{sec:conclusion}
We have presented FloodTrace, a web-based application for quick and accurate annotation of flooded regions that better enables crowdsourcing. This framework was built directly from requirements gathered from researchers in flood extent mapping. Our application utilizes a region's elevation data to provide users insight in 3D and guide users with semi-automatic annotation tools. This application has already been utilized for the creation of training data by domain experts for cutting-edge flood extent mapping models \cite{sami2024evanetelevationguidedfloodextent}. Our application brings to the web environment, traditionally complex segmentation methods using topological data analysis to greatly outperform the state-of-the-art elevation-guided tool in efficiency. Our experimental user study shows that the benefits of our tools can be found even for untrained users. With our method for aggregating and correcting groups of annotations, researchers are able to use crowdsourced annotations as training data for their ML models with equal performance to fully expert-labeled data. This improves their workflows greatly, replacing an extremely time-consuming annotation process with a much quicker correction step. 


Our work does have limitations. Our work only supports labeling for two classes of pixels, flooded or dry. Selecting features of the data with more specific labels, such as permanent water bodies, is useful for some flood detection models, and future work could tackle extending elevation-guided annotation tools to tasks with more informative labels.
In addition, there are limitations with regards to our use of DEM data for our evaluation. The use of DEM data means that objects above the ground level are absent. For some regions, such as heavily urban environments, the presence of these objects could affect water propagation enough to cause the semi-automatic tools to label things incorrectly, for example causing the BFS to falsely label downstream pixels as flooded when the water actually stopped at a building. This limitation could be addressed by using digital surface model (DSM) data or models built from LiDAR data, which take into account above-ground structures. Our application can support any form of elevation data as long as it corresponds to the input satellite imagery, but as of yet we have only evaluated the use of DEM data. Experimentation with other elevation models is left for future work.
Finally, as with the machine learning algorithms that rely on elevation data, our system requires that the elevation data and flooded aerial imagery being annotated are collected at similar dates. If not, deviations between elevation at the time of flooding and time of data collection could lead to error. 

\acknowledgments{
This work was partly funded by NSF Collaborative Research Awards 2401274, 2221812, and 2106461, and NSF PPoSS Planning and Large awards 2217036 and 2316157.
}

\bibliographystyle{abbrv-doi}
\bibliography{heightmap-annotation.bib}

\end{document}